\documentclass[11pt]{article}
\usepackage[utf8]{inputenc}
\usepackage[T1]{fontenc}
\usepackage{graphicx}
\usepackage[colorlinks=true,linkcolor=blue,citecolor=blue,urlcolor=blue]{hyperref}
\usepackage[numbers]{natbib}
\usepackage{amsmath}
\usepackage{setspace}
\usepackage{geometry}
\title{LLM Agent Swarm for Hypothesis-Driven Drug Discovery}

\author{%
  Kevin Song\thanks{Corresponding author: \href{mailto:kmsong@uab.edu}{kmsong@uab.edu}}\textsuperscript{1,2}, %
  Andrew Trotter\textsuperscript{1}, %
  Jake Y.\ Chen, PhD\textsuperscript{1,2}%
  \\[1ex]
  \textsuperscript{1}Systems Pharmacology AI Research Center, The University of Alabama at Birmingham\\
  \textsuperscript{2}Department of Biomedical Engineering, The University of Alabama at Birmingham%
}

\begin{document}
\maketitle

\begin{abstract}
Drug discovery remains a formidable challenge: more than 90 percent of candidate molecules fail in clinical evaluation, and development costs often exceed one billion dollars per approved therapy. Disparate data streams, from genomics and transcriptomics to chemical libraries and clinical records, hinder coherent mechanistic insight and slow progress. Meanwhile, large‐language models excel at reasoning and tool integration but lack the modular specialization and iterative memory required for regulated, hypothesis‐driven workflows. We introduce PharmaSwarm, a unified multi‐agent framework that orchestrates specialized LLM “agents” to propose, validate, and refine hypotheses for novel drug targets and lead compounds. Each agent accesses dedicated functionality—automated genomic and expression analysis; a curated biomedical knowledge graph; pathway enrichment and network simulation; interpretable binding affinity prediction—while a central Evaluator LLM continuously ranks proposals by biological plausibility, novelty, in silico efficacy, and safety. A shared memory layer captures validated insights and fine‐tunes underlying submodels over time, yielding a self‐improving system. Deployable on low‐code platforms or Kubernetes‐based microservices, PharmaSwarm supports literature‐driven discovery, omics‐guided target identification, and market‐informed repurposing. We also describe a rigorous four‐tier validation pipeline spanning retrospective benchmarking, independent computational assays, experimental testing, and expert user studies to ensure transparency, reproducibility, and real‐world impact. By acting as an AI copilot, PharmaSwarm can accelerate translational research and deliver high‐confidence hypotheses more efficiently than traditional pipelines.
\end{abstract}

\newpage

\doublespacing

\section{Introduction}
Drug development is notorious for high attrition rates and staggering costs. Over ninety percent of candidate molecules fail during clinical evaluation, leading to an average expenditure of one to two billion dollars per approved therapeutic \cite{Waring2015}. These failures reflect gaps in the ability to integrate genomic, chemical, and clinical data into coherent mechanistic hypotheses. Traditional pipelines process each data type in isolation, resulting in delayed feedback loops and missed opportunities for cross-domain synergy \cite{Pushpakom2019}.

Large language models such as GPT-4 and Gemini 2.5 Pro have demonstrated remarkable capabilities in text comprehension, reasoning, and integration with external tools \cite{OpenAI2023, Bommasani2021}. Nonetheless, single-model deployments lack the specialization, memory retention, and mechanistic rigor required for regulated biomedical workflows. Recent work has shown that multi-agent LLM systems can outperform monolithic models in tasks ranging from retrosynthetic chemistry \cite{Segler2020} to autonomous laboratory protocol generation \cite{Pompe2023} and ecological forecasting \cite{Nguyen2023}. These systems exploit heterogeneous reasoning strategies and iterative feedback to explore vast hypothesis spaces more effectively \cite{Li2022, Wang2023}.

Building on these advances, we introduce PharmaSwarm, a competitive, interpretable, and continuously evolving swarm of LLM agents tailored for hypothesis-driven drug discovery. PharmaSwarm unifies multi-modal data ingestion, mechanistic simulation, and transparent decision-making. A central Evaluator LLM ranks and critiques proposals, while a shared memory store of graph updates and vector embeddings enables continuous learning and submodel fine-tuning. This paper describes the overall architecture (Section 2), the iterative swarm workflow (Section 3), illustrative use cases (Section 4), and proposed validation strategies (Section 5). We conclude with deployment considerations and future extensions (Section 6).

\section{System Architecture}
PharmaSwarm is designed as a three‐layered architecture, each layer handling a distinct set of responsibilities yet tightly integrated to ensure seamless data flow and iterative refinement.

\begin{figure}[ht]
  \centering
  \includegraphics[width=1\textwidth]{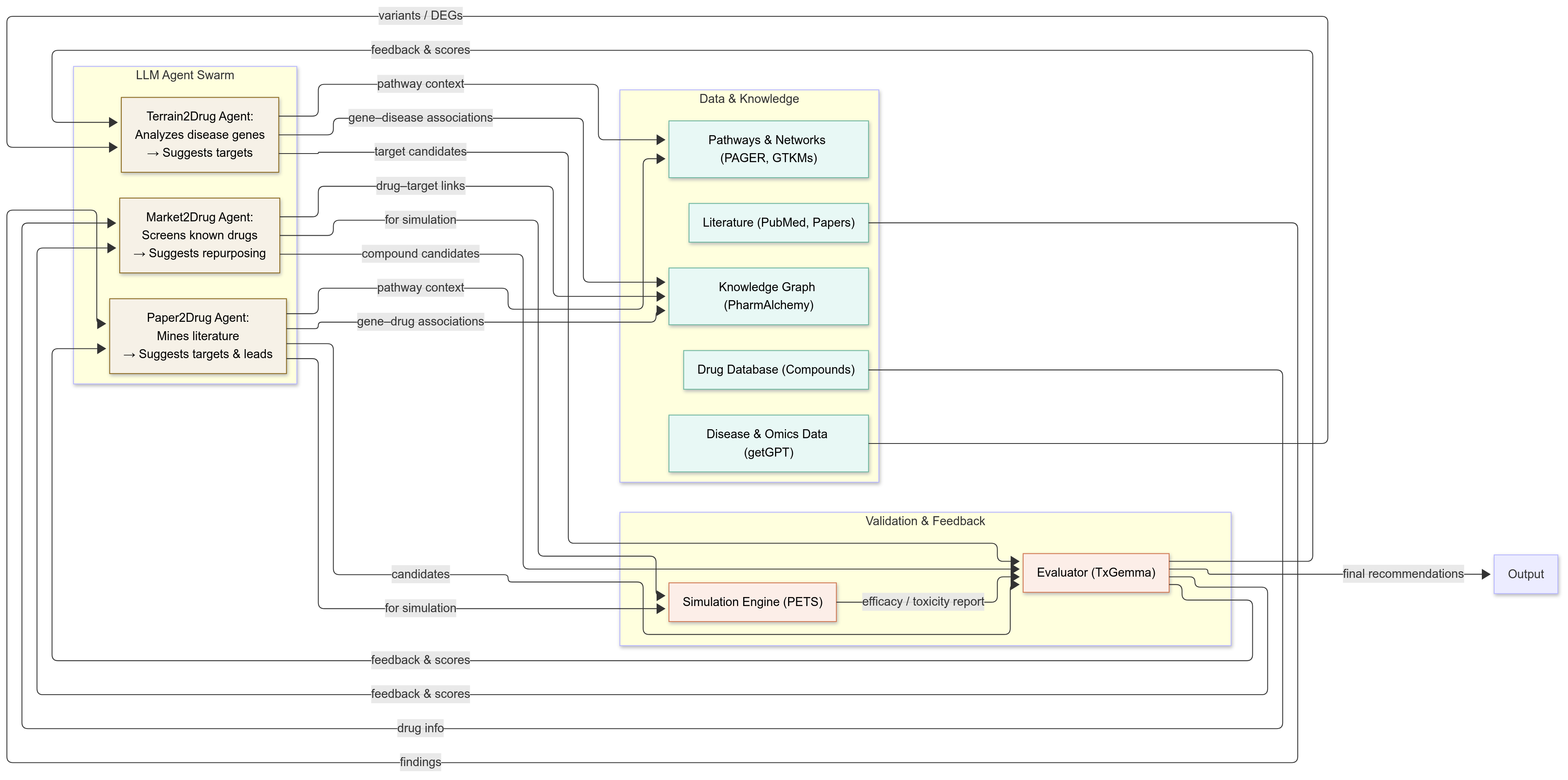}
  \caption{\textbf{LLM Agent Swarm Architecture.} A modular, agent-based pipeline integrates heterogeneous biomedical knowledge—pathway and network databases, literature corpora, a unified knowledge graph and compound repositories—with three specialized LLM agents (Terrain2Drug, Market2Drug and Paper2Drug) that propose disease targets and candidate compounds. Proposals are subjected to in silico pharmacological simulations (PETS) and efficacy/toxicity scoring by a dedicated evaluator, and bidirectional feedback loops continuously enrich both the shared knowledge base and subsequent agent outputs, yielding interpretable therapeutic hypotheses through iterative refinement.}
  \label{fig:arch}\label{fig:arch}
\end{figure}

\subsection*{Data \& Knowledge Layer}
The foundation of PharmaSwarm is built on comprehensive ingestion and preprocessing of diverse biomedical data. The getGPT module extracts G.E.T. lists of disease-related genetic variants, differentially expressed genes (DEGs), and drug targets by interfacing with the Gene Expression Omnibus and the Open Targets and OpenTarget Genetics APIs to retrieve known drug targets, genome‐wide association study (GWAS) loci, fine‐mapped variants, and gene‐trait association scores. Parallel pipelines query the Gene Expression Omnibus (GEO) for transcriptomic datasets, performing normalization and differential expression analysis. Scientific literature is harvested via PubMed and bioRxiv APIs, where title‐abstract pairs and open‐access full texts are parsed for downstream LLM summarization and gene set curation. Chemical and pharmacological knowledge is sourced from ChEMBL, DrugBank, and the PharmAlchemy knowledge base (i.e., a "database of databases" of drugs-genes-diseases), while pathway context is provided by KEGG and Reactome REST services. The PAGER API supplies network‐based gene sets with enrichment scores, GeneTerrain Knowledge Maps (GTKMs) render spatial topography of genetic interaction and expression networks, and visualization of on- and off-target effects from simulation results in the Validation \& Evaluation layer.

\subsection*{LLM Agent Swarm Layer}
At the core of PharmaSwarm are three specialized agents, each implemented as containerized microservices with access to the shared knowledge base:
\begin{itemize}
  \item \textbf{Terrain2Drug Agent} focuses on omics‐driven discovery. It accepts seed gene lists from getGPT and PAGER, projects them onto raw GTKMs, and identifies high‐degree network hubs via graph‐topology algorithms. These hubs become candidate targets, with pathway enrichment statistics guiding prioritization.
  \item \textbf{Paper2Drug Agent} conducts automated literature mining. Using LLM‐templated prompts, it extracts explicit and implicit target–compound relationships from text, then validates these through multi‐hop traversals in the PharmAlchemy knowledge graph to ensure mechanistic consistency.
  \item \textbf{Market2Drug Agent} synthesizes market and community intelligence. It streams regulatory bulletins (e.g.\ FDA notices), clinical‐trial registry updates, financial APIs (e.g.\ Bloomberg), and social‐media sentiment (Twitter, Reddit) into semantic embeddings. These signals flag compounds with emerging clinical relevance, which are then cross‐matched to chemical similarity scores and network impact predictions.
\end{itemize}
Each agent maintains a local prompt‐history buffer and accesses a shared embedding store (via a vector database) to capture inter‐agent knowledge and avoid redundant explorations.

\subsection*{Validation \& Evaluation Layer}
\begin{figure}[ht]
\centering
\includegraphics[width=0.7\textwidth]{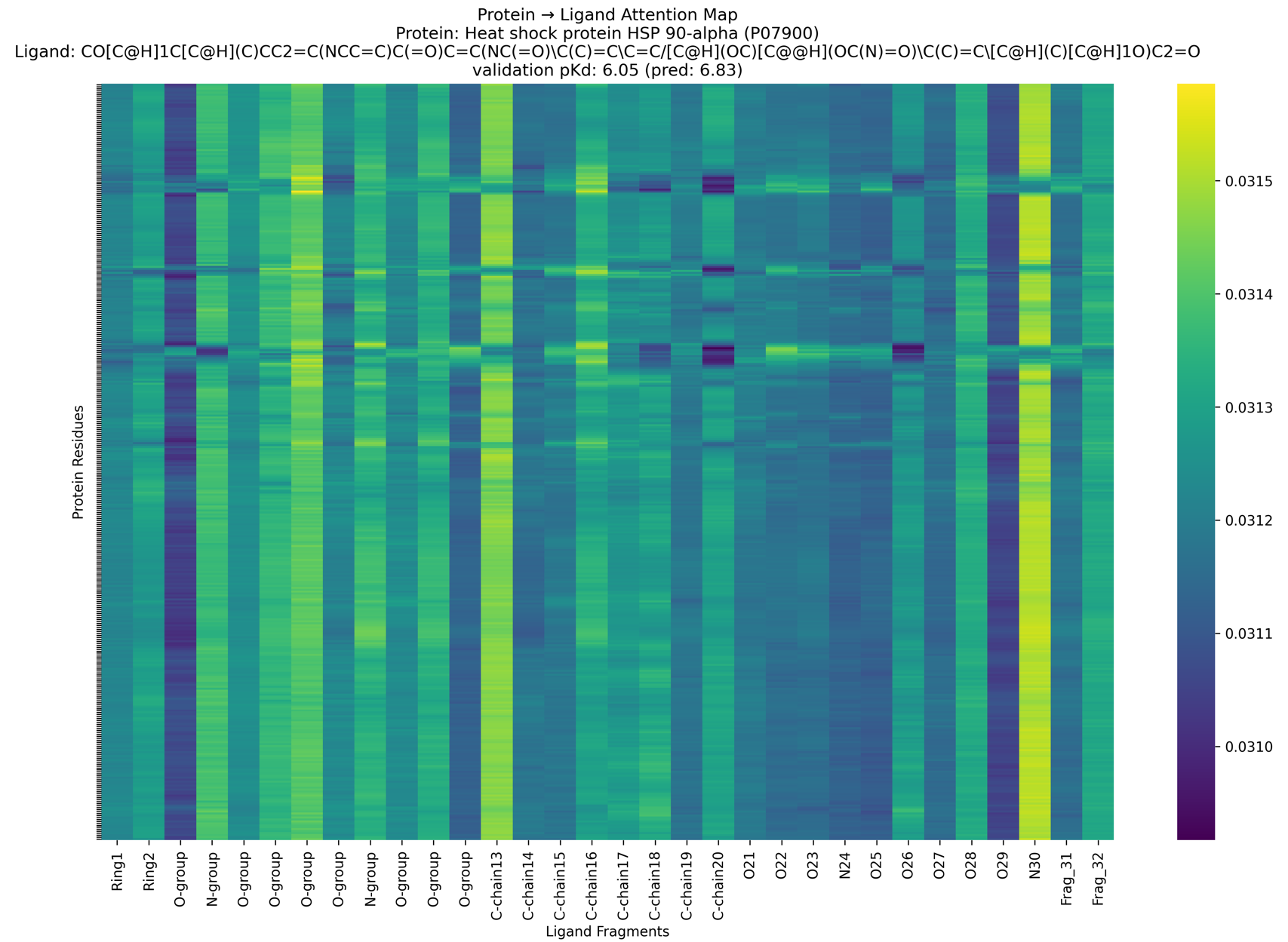}
\caption{\textbf{iBAM for the HSP90-alpha–Ligand Complex.} This heatmap visualizes the cross-attention weights between the protein residues of HSP90-alpha (UniProt ID P07900) and the substructures of a candidate ligand (shown as Morgan fingerprint fragments). Warmer (yellow) regions correspond to higher attention, highlighting critical residue–substructure interactions that drive the predicted binding affinity. Here, the predicted pKd is 6.83, while the experimentally determined pKd is 6.05. By providing a clear, interpretable view of how the model “focuses” on specific protein–ligand contacts, iBAM offers a valuable tool for guiding rational drug design and facilitating targeted lead optimization in structure-based drug discovery.}
\label{fig:ibam}
\end{figure}

Once agents propose targets or compounds, the Validation \& Evaluation layer quantifies mechanistic plausibility and safety:
\begin{itemize}
  \item \textbf{Pharmacological Efficacy and Toxicity Simulation (PETS) Engine} executes multi‐scale network propagation of compound perturbations across tissue‐specific protein–protein interaction networks to yield standardized efficacy and toxicity scores \cite{Song2025PETS}.
  \item \textbf{Interpretable Binding Affinity Map (iBAM) Module} employs a cross‐attention architecture between ESM2 protein embeddings and ChemBERTa molecular embeddings, producing both affinity estimates and structure-free residue–chemical substructure attention maps for transparent SAR analysis and lead optimization (\hyperref[fig:ibam]{Figure 2}).
  \item \textbf{Central Evaluator} is a TxGemma-powered LLM instance that ingests agent proposals, simulation outputs, and binding maps \cite{wang2025txgemmaefficientagenticllms}. It applies a multi‐criteria scoring rubric—data support, mechanistic coherence, novelty, safety margin, and interpretability—and generates actionable feedback to each agent, closing the loop for iterative refinement.
\end{itemize}

\subsection*{Orchestration \& Deployment}
PharmaSwarm’s layers are orchestrated via flexible workflow engines or container platforms:
\begin{itemize}
  \item \textbf{Low-Code Platforms}: n8n, Apache Airflow, or Prefect for visual DAG construction, built‐in retry logic, and monitoring dashboards.
  \item \textbf{Kubernetes Microservices}: Argo Workflows or Kubeflow Pipelines for scalable, cloud‐native execution, leveraging Helm charts for reproducible deployments and Istio for secure service‐mesh communications.
  \item \textbf{CI/CD Integration}: GitOps pipelines ensure automatic redeployment of updated agent code or model weights, with Canary releases and A/B testing to validate system-wide performance improvements.
\end{itemize}

\section{Iterative Swarm Workflow}
PharmaSwarm operates as a closed‐loop, iterative process in which user input, agent-based hypothesis generation, mechanistic validation and evaluative feedback drive continuous refinement. At the start of each cycle, the end user specifies the disease or biological context of interest and may impose constraints such as preferred target classes, compound libraries, or acceptable safety thresholds. The orchestrator—implemented on a workflow engine such as n8n or as Kubernetes-native microservices—then invokes getGPT and PAGER to retrieve and rank relevant genetic variants and expression signatures, producing a seed gene list with associated statistical annotations.

Simultaneously, the orchestrator spins up the three agent services in parallel. The Terrain2Drug Agent consumes the seed list and projects it onto the GeneTerrain Knowledge Maps to identify high-degree network hubs. The Paper2Drug Agent submits literature retrieval queries to PubMed and bioRxiv, using LLM-driven extraction to propose target–compound pairs grounded in recent publications. The Market2Drug Agent ingests regulatory notices, clinical-trial updates, financial news feeds and social media signals, computing chemical similarity and network impact scores to highlight repurposing candidates. All inter-service communication is handled via a message queue, ensuring asynchronous scalability, and each agent logs its raw proposals—targets or compounds—into the shared memory store.

Once the initial hypotheses are collected, the Validation and Evaluation layer is engaged. The PETS Simulation Engine receives candidate compounds for simulating in silico perturbation of tissue-specific protein interaction networks, generating standardized efficacy and toxicity metrics. In parallel, the interpretable binding affinity module computes attention‐based binding maps and pK\textsubscript{d} estimates for each target–ligand pair.

All simulation outputs, along with provenance metadata linking back to graph traversals, literature citations and input data versions, are aggregated for the central Evaluator LLM. This Evaluator applies a multi-criteria rubric—assessing empirical support, mechanistic coherence, novelty relative to existing knowledge, predicted safety margins, and human-readable interpretability—and assigns each hypothesis a composite score.

Detailed feedback is formulated as structured prompts, which are then routed back to the individual agents. For example, the Terrain2Drug Agent may be instructed to deprioritize hubs with low druggability or to explore alternative network modules; the Paper2Drug Agent may receive guidance on refining search terms to focus on kinase inhibitors; and the Market2Drug Agent may be directed to incorporate additional social-media sentiment filters. Agents update their internal prompt histories and heuristics accordingly, and the orchestrator initiates the next iteration.

Cycles continue until convergence criteria—such as reaching a maximum iteration count, achieving a minimum improvement threshold in top-ranked scores, or exhausting the search space—are satisfied. The workflow concludes by exporting a final prioritized list of targets and compounds, complete with interactive provenance reports, simulation dashboards, and binding maps to support expert decision-making and downstream experimental validation.

\begin{figure}[ht]
  \centering
  \includegraphics[width=1\textwidth]{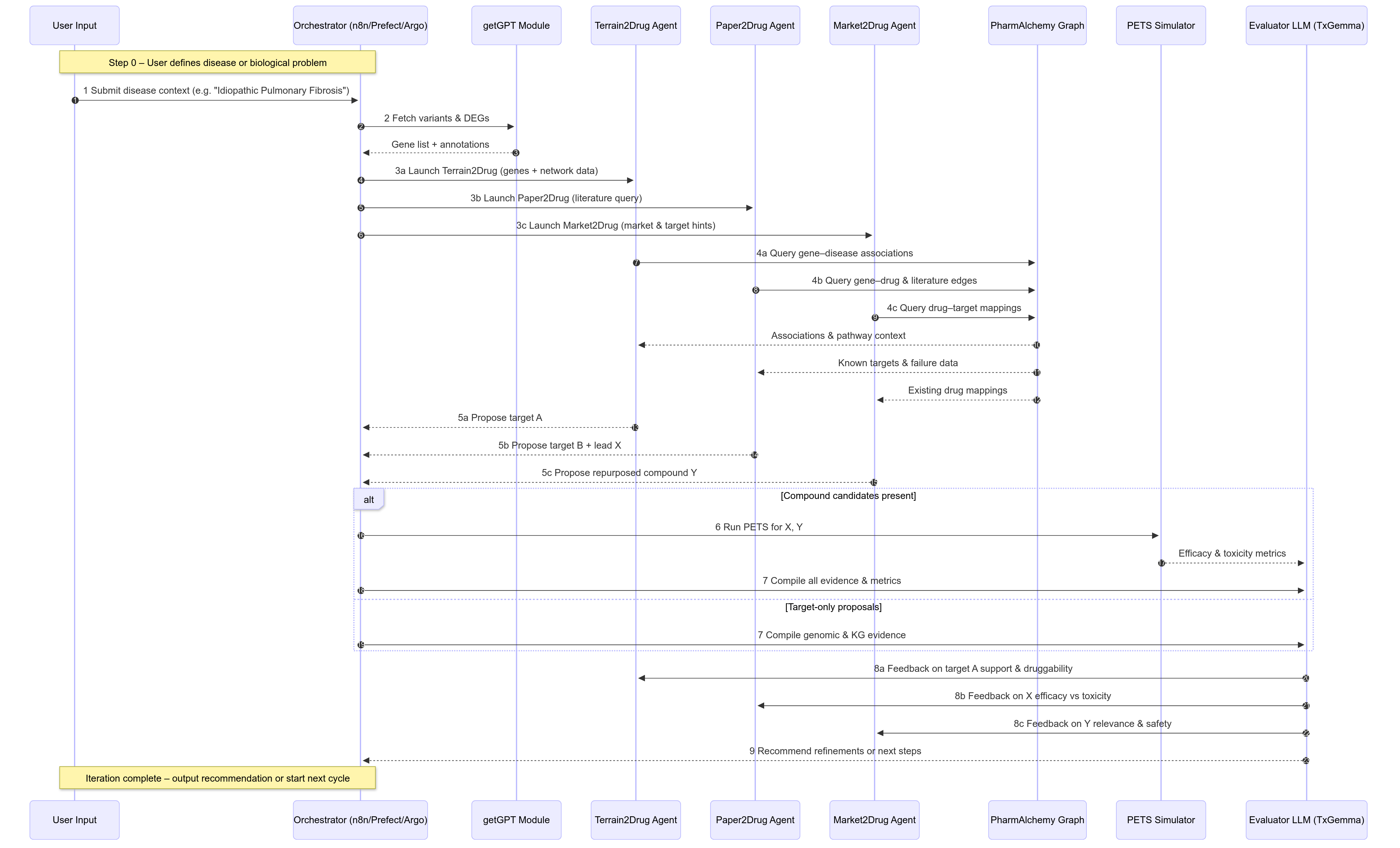}
  \caption{\textbf{Iterative Workflow and Example Scenario.} The PharmaSwarm workflow proceeds in iterative cycles of hypothesis generation and refinement. Here we describe a typical cycle, then illustrate how it manifests in three example scenarios (Paper2Drug, Terrain2Drug, and Market2Drug), which correspond to different initial strategies the swarm can take. This figure outlines a simplified sequence of interactions in one iteration of the swarm’s workflow for a given disease input.}
  \label{fig:seq}
\end{figure}

\section{Example Use‐Case Scenarios}

\paragraph{Paper2Drug.}  
In the Paper2Drug scenario, PharmaSwarm agents perform end‐to‐end literature mining to surface novel target–compound hypotheses directly from scientific publications. First, the orchestrator issues API calls to PubMed and bioRxiv, retrieving both abstracts and, when available, open‐access full texts. The Paper2Drug Agent then applies a chain‐of‐thought prompting protocol in GPT‐4 or Gemini, asking the model to identify explicit mentions of drug–target pairs, as well as implicit mechanistic clues—such as pathway modulations or phenotype rescue experiments—that suggest novel interactions. Extracted candidate pairs are annotated with metadata (publication DOI, sentence index, confidence score) and passed to PharmAlchemy for graph‐based cross‐validation: multi‐hop traversals confirm whether the proposed targets connect to disease nodes via known druggable pathways, and whether similar scaffold–target pairs have precedent. Promising candidates are forwarded to the PETS Simulation Engine, where in silico perturbation assesses efficacy (pathway reversion) and off‐target toxicity. Finally, the agent compiles binding‐attention maps, highlighting residue–substructure alignments that explain predicted affinities, and assembles a provenance report linking each hypothesis to specific publications, graph paths, and simulation metrics. This end‐to‐end pipeline transforms unstructured text into mechanistically grounded, experimentally testable drug hypotheses.

\paragraph{Terrain2Drug.}  
Terrain2Drug harnesses patient‐derived omics to reveal high‐impact regulatory hubs and their modulators. Starting with bulk RNA‐seq datasets from GEO or single‐cell profiles from TCGA, the orchestrator standardizes raw counts through established pipelines (e.g.\ DESeq2 for bulk, Seurat for single‐cell) and computes differential expression against relevant controls. The resulting gene lists are projected onto GeneTerrain Knowledge Maps, which render expression changes as topographical elevations within a tissue‐specific protein–protein interaction network. Peaks—genes with high fold‐change and centrality—are flagged as candidate regulators. Next, the agent queries KEGG, Reactome, and the PAGER API to perform both over‐representation and topology‐aware enrichment, identifying pathways and upstream nodes that drive disease phenotypes. PharmAlchemy traversals then map these hubs to compounds with known or predicted binding profiles. Each hub–compound pair undergoes PETS simulation, evaluating network resilience and PK/PD dynamics, alongside binding‐affinity attention mapping. Validated pairs are stored in shared memory, which updates the TxGemma submodel, sharpening its heuristics for subsequent cycles. This iterative process refines both target selection and compound optimization in a data‐driven, mechanistically transparent manner.

\paragraph{Market2Drug.}  
Market2Drug integrates real‐world intelligence to prioritize repurposing opportunities from existing pharmacopeias. The orchestrator streams structured and unstructured market signals: FDA approval bulletins via OpenFDA APIs, clinical‐trial registry updates from ClinicalTrials.gov, financial data from Bloomberg or Refinitiv, and social‐media discussions on Twitter and Reddit (subreddits like \texttt{r/medicine}, \texttt{r/pharmacology}). A specialized LLM pipeline converts these heterogeneous inputs into normalized sentiment and relevance scores, identifying compounds that are gaining traction or facing safety alerts. In parallel, the agent computes chemical similarity metrics using ChemBERTa embeddings against a library of approved drugs, and submits top candidates to PETS for network‐impact simulations. PharmAlchemy graph queries verify mechanistic links between repurposing candidates and the disease context, ensuring each recommendation is grounded in biological plausibility. Social‐media sentiment and regulatory flags are then combined with efficacy and toxicity scores to rank compounds by a composite “clinical‐actionability” metric. The resulting prioritized list reflects both scientific rigor and market readiness, making Market2Drug particularly suited for rapid translation and real‐world validation.

\section{Validation Strategies}

Establishing confidence in PharmaSwarm’s outputs requires a rigorous, four‐tiered validation pipeline that progresses from historical reconstruction to real‐world expert feedback, each evaluated with standardized metrics.

\subsection*{Tier 1: Retrospective Benchmarking}
We first reconstruct classic discovery cases (e.g.\ idiopathic pulmonary fibrosis, triple‐negative breast cancer) using only the data available at the time—GWAS summaries, expression datasets, literature corpora, and compound libraries. Running PharmaSwarm in a “frozen” mode (no subsequent fine‐tuning), we compare its top‐K target and compound rankings against the clinical “ground truth.” Performance is quantified by Recall@K and Precision@K for known leads, Kendall’s Tau rank correlation with actual validation order, and Mean Average Precision (MAP) across indications. Bootstrapped confidence intervals assess significance relative to baseline methods (e.g.\ a single‐agent LLM or standard network‐based repurposing).

\subsection*{Tier 2: Prospective In Silico Assessment}
Novel hypotheses from Tier~1 undergo independent computational testing. We perform both blind and targeted molecular docking (AutoDock Vina, Glide) to obtain binding energies and validate poses, followed by short (50--100\,ns) molecular dynamics simulations to assess complex stability (root-mean-square deviation, hydrogen-bond occupancy). Predicted ADMET profiles are generated with pkCSM or ADMETlab, and PETS simulations are rerun on alternative PPI networks to verify the robustness of efficacy and toxicity scores. Finally, we correlate PharmaSwarm’s predicted $pK_d$ and efficacy metrics with docking affinities (Pearson’s $r$), MD stability, and known drug-label ADMET flags to provide quantitative validation.

\subsection*{Tier 3: Experimental Evaluation}
Top candidates advance to laboratory assays, where binding affinities ($K_d$) are measured by SPR or ITC, and cellular phenotypic screens yield $IC_{50}$ values and pathway‐specific readouts (e.g.\ reporter assays). Off‐target activity is profiled using receptor or kinase panels (Cerep, DiscoverX), and in vivo pilot studies in rodent models (fibrosis, xenograft) assess pharmacokinetics, efficacy biomarkers, and safety endpoints. Predefined go/no–go criteria—such as $K_d<100\,$nM, $IC_{50}<1\,\mu$M, and acceptable therapeutic index—determine hit rates and progression.

\subsection*{Tier 4: Expert User Studies}
To measure practical utility, we engage medicinal chemists, pharmacologists, and translational biologists in timed comparisons between PharmaSwarm‐guided and conventional workflows. Participants generate and document top‐N hypotheses under each approach. We track time‐to‐hypothesis, collect expert ratings on mechanistic plausibility and novelty (Likert scale), and survey decision confidence before and after. Paired tests (e.g.\ Wilcoxon signed‐rank) quantify improvements in speed, quality, and confidence.

\section{Discussion and Future Directions}

PharmaSwarm demonstrates how a modular, multi‐agent approach can integrate diverse data modalities, mechanistic simulations, and interpretable AI into a single, coherent pipeline. By containerizing each agent and orchestrating workflows through either low‐code platforms or Kubernetes‐native frameworks, the system achieves both reproducibility and elastic scalability—from a single workstation to high‐performance cloud clusters. Detailed provenance tracking at every step addresses regulatory and audit requirements, ensuring that each recommendation is accompanied by traceable evidence: knowledge graph traversals, literature citations, simulation logs, and binding‐attention maps.

Nonetheless, several challenges and opportunities remain. First, input data quality and completeness heavily influence hypothesis generation; integrating data quality metrics and uncertainty quantification at the ingestion stage will improve robustness. Second, while shared memory enables continuous learning, model drift and catastrophic forgetting must be managed through targeted fine‐tuning schedules and checkpointing strategies. Third, federated learning could extend PharmaSwarm to private or proprietary datasets—allowing multiple institutions to collaboratively improve submodels without sharing raw data. Fourth, the integration of emerging functional genomics assays—such as CRISPR screening and single‐cell multi‐omics—will provide higher‐resolution insights into cell‐type specific targets and drug responses. Fifth, real‐time ingestion of preprints, clinical trial updates, and regulatory notices will keep the system aligned with the latest developments, enhancing its responsiveness in rapidly evolving fields such as oncology or infectious disease.

Looking further ahead, incorporating clinical‐outcome prediction modules—trained on electronic health record cohorts or trial results—could close the loop between in silico hypothesis generation and patient impact, enabling end‐to‐end AI‐driven translational research. Embedding human‐in‐the‐loop review interfaces and visual analytics will foster adoption among domain experts, balancing automated suggestion with expert judgment. Finally, rigorous benchmarking against evolving gold‐standard datasets and participation in community challenges will drive continuous improvement and transparency.

\section{Conclusion}
PharmaSwarm presents a novel, competitive swarm of LLM agents that unifies heterogeneous biomedical data, mechanistic simulation engines, and interpretable machine learning into a continuously evolving framework for hypothesis-driven drug discovery. Its layered architecture—spanning data ingestion, specialized agent reasoning, and rigorous validation—provides a transparent and auditable path from raw data to prioritized targets and compounds. By combining retrospective and prospective validation strategies with flexible deployment options and a shared memory of validated insights, PharmaSwarm not only accelerates the generation of actionable hypotheses but also adapts over time to new knowledge and emerging technologies. As translational pipelines increasingly rely on AI co-pilots, PharmaSwarm offers a blueprint for how multi-agent systems can deliver scientific rigor, regulatory compliance, and real-world impact in the quest for next-generation therapeutics.

\bibliographystyle{unsrtnat}

\begin{thebibliography}{99}

\bibitem{Waring2015}
Waring MJ, Arrowsmith J, Leach AR, et~al. An analysis of the attrition of drug candidates from four major pharmaceutical companies. \emph{Nature Reviews Drug Discovery}. 2015;14(7):475--486.

\bibitem{Pushpakom2019}
Pushpakom S, Iorio F, Eyers PA, et~al. Drug repurposing: progress, challenges and recommendations. \emph{Nature Reviews Drug Discovery}. 2019;18:41--58.

\bibitem{OpenAI2023}
OpenAI. GPT-4 Technical Report. arXiv:2303.08774. 2023.

\bibitem{Bommasani2021}
Bommasani R, Hudson D, Adeli E, et~al. On the Opportunities and Risks of Foundation Models. arXiv:2108.07258. 2021.

\bibitem{Segler2020}
Segler MHS, Preuss M, Waller M. Planning chemical syntheses with deep neural networks and symbolic AI. \emph{Nature}. 2020;555:604--610.

\bibitem{Pompe2023}
Pompe N, et~al. Autonomous laboratory planning with multi-agent LLM systems. \emph{Nature Communications}. 2023;14:2345.

\bibitem{Nguyen2023}
Nguyen T, et~al. Agent-based climate modeling with language models. \emph{Geophysical Research Letters}. 2023;50(10):e2022GL100123.

\bibitem{Li2022}
Li X, Gan Z, Carin L. Beyond single-agent LLMs: A survey on multi-agent LLM interactions. arXiv:2211.00845. 2022.

\bibitem{Wang2023}
Wang Y, Jiang L, Zhou M, Smith NA. Evaluating multi-agent collaboration in large language models. In: \emph{Proceedings of the 61st Annual Meeting of the Association for Computational Linguistics (ACL)}. 2023:2334--2346.

\bibitem{wang2025txgemmaefficientagenticllms}
Wang E, Schmidgall S, Jaeger PF, Zhang F, Pilgrim R, Matias Y, Barral J, Fleet D, Azizi S. TxGemma: Efficient and Agentic LLMs for Therapeutics. arXiv:2504.06196. 2025.

\bibitem{Ochoa2023}
Ochoa D, Hercules A, Carmona M, et~al. The Open Targets Platform: Re-imagining drug target identification. \emph{Nucleic Acids Research}. 2023;51(D1):D1305--D1316.

\bibitem{Edgar2002}
Edgar R, Domrachev M, Lash AE. Gene Expression Omnibus: NCBI gene expression and hybridization array data repository. \emph{Nucleic Acids Research}. 2002;30(1):207--210.

\bibitem{Gaulton2017}
Gaulton A, Hersey A, Nowotka M, et~al. The ChEMBL database in 2017. \emph{Nucleic Acids Research}. 2017;45(D1):D945--D954.

\bibitem{Wishart2018}
Wishart DS, et~al. DrugBank 5.0: a major update to the DrugBank database for 2018. \emph{Nucleic Acids Research}. 2018;46(D1):D1074--D1082.

\bibitem{Sosa2020}
Sosa DN, Derry A, Guo M, Wei E, Brinton C, Altman RB. A literature-based knowledge graph embedding method for identifying drug repurposing opportunities in rare diseases. \emph{Pacific Symposium on Biocomputing}. 2020:589--600.

\bibitem{Kanehisa2000}
Kanehisa M, Goto S. KEGG: Kyoto Encyclopedia of Genes and Genomes. \emph{Nucleic Acids Research}. 2000;28(1):27--30.

\bibitem{Jassal2020}
Jassal B, Matthews L, Viteri G, et~al. The Reactome pathway knowledgebase. \emph{Nucleic Acids Research}. 2020;48(D1):D498--D503.

\bibitem{Song2025PETS}
Song K, Chen JY. Integrative multi-scale network simulation for precision drug repurposing with PETS. \emph{bioRxiv}. 2025.

\bibitem{Rives2021}
Rives A, Meier J, Sercu T, et~al. Biological structure and function emerge from scaling unsupervised learning to 250 million protein sequences. \emph{Proceedings of the National Academy of Sciences}. 2021;118(15):e2016239118.

\bibitem{Chithrananda2022}
Chithrananda S, Grand G, Ramsundar B. ChemBERTa: Large-scale self-supervised pretraining for molecular property prediction. arXiv:1910.06725. 2022.

\bibitem{Airflow2017}
Apache Airflow. 2017. \url{https://airflow.apache.org/}

\bibitem{Prefect2023}
Prefect. 2023. \url{https://www.prefect.io/}

\bibitem{DiTommaso2017}
Di Tommaso P, et~al. Nextflow enables reproducible computational workflows. \emph{Nature Biotechnology}. 2017;35(4):316--319.

\bibitem{Koster2012}
Köster J, Rahmann S. Snakemake—a scalable bioinformatics workflow engine. \emph{Bioinformatics}. 2012;28(19):2520--2522.

\bibitem{Kubeflow2023}
Kubeflow Pipelines. 2023. \url{https://www.kubeflow.org/}

\bibitem{Argo2024}
Argo Workflows. 2024. \url{https://argoproj.github.io/argo-workflows/}

\bibitem{n8n2024}
n8n. 2024. \url{https://n8n.io/}

\end{thebibliography}

\end{document}